\documentclass[runningheads]{llncs}
\usepackage{inconsolata}
\usepackage{amsmath}
\usepackage{graphicx}
\usepackage{multirow}
\usepackage{latexsym}
\usepackage{amssymb}
\usepackage{makecell}
\usepackage{booktabs}
\usepackage{bbm}
\usepackage{wrapfig}
\usepackage{lipsum} 
\usepackage{xcolor}
\usepackage{caption}

\definecolor{Mycolor1}{HTML}{A01D14}
\definecolor{Mycolor2}{HTML}{DBB15B}
\definecolor{Mycolor3}{HTML}{253D8C}
\definecolor{Mycolor4}{HTML}{256b48}

\begin{document}
\title{Editing Personality For Large Language Models
}
\author{
Shengyu Mao\inst{1},
Xiaohan Wang\inst{1},
Mengru Wang\inst{1},
Yong Jiang\inst{2},
Pengjun Xie\inst{2},
Fei Huang\inst{2},
Ningyu Zhang\inst{1\dag}
}
\authorrunning{Shengyu Mao, et al.}

\institute{Zhejiang University \and
Alibaba Group \\
\email{\{shengyu, zhangningyu\}@zju.edu.cn}}
\maketitle            
\begin{abstract}
This paper introduces an innovative task focused on editing the personality traits of Large Language Models (LLMs). This task seeks to adjust the models' responses to opinion-related questions on specified topics since an individual's personality often manifests in the form of their expressed opinions, thereby showcasing different personality traits. Specifically, we construct \textbf{PersonalityEdit}, a new benchmark dataset to address this task. Drawing on the theory in Social Psychology \cite{goldberg1990alternative}, we isolate three representative traits, namely \textsc{Neuroticism}, \textsc{Extraversion}, and \textsc{Agreeableness}, as the foundation for our benchmark. We then gather data using GPT-4, generating responses that align with a specified topic and embody the targeted personality trait. We conduct comprehensive experiments involving various baselines and discuss the representation of personality behavior in LLMs. Our findings uncover potential challenges of the proposed task, illustrating several remaining issues. We anticipate that our work can stimulate further annotation in model editing and personality-related research.
\footnotetext{\dag~Corresponding Author}
\footnotetext{Code is available at \url{https://github.com/zjunlp/EasyEdit}}

\end{abstract}

\section{Introduction}
Large Language Models (LLMs) have made remarkable strides in modeling language distributions and excelling in a wide array of NLP tasks \cite{openai2023gpt4,DBLP:journals/corr/abs-2305-08732,DBLP:journals/corr/abs-2303-18223}. 
More recent studies \cite{DBLP:journals/corr/abs-2304-03442,DBLP:journals/corr/abs-2305-16867,agent_survey} expand our understanding of LLMs in role-playing scenarios, which effectively serve as a rich array of agents, embodying a multitude of potential characters within an expansive multiverse \cite{DBLP:journals/corr/abs-2305-16367}.

Unlike LLMs, humans exhibit distinct personalities, and each person has a certain degree of personality in their response to events and actions \cite{goldberg1981language}.
The remarkable role-playing capabilities of LLMs have promoted the investigation of their personality~\cite{do_llms_posess_a_personality,personality_traits_in_llms,MPI}.
Meanwhile, recent works have been attempting to edit the topic-level knowledge in LLMs~\cite{serac,MEND,rome},
this leads us to the research question: \textbf{Can we edit the personality for LLMs?}
Note that implementing editing methods for LLMs' personality can:
\begin{figure}
    \centering
    \includegraphics[width=1\textwidth]{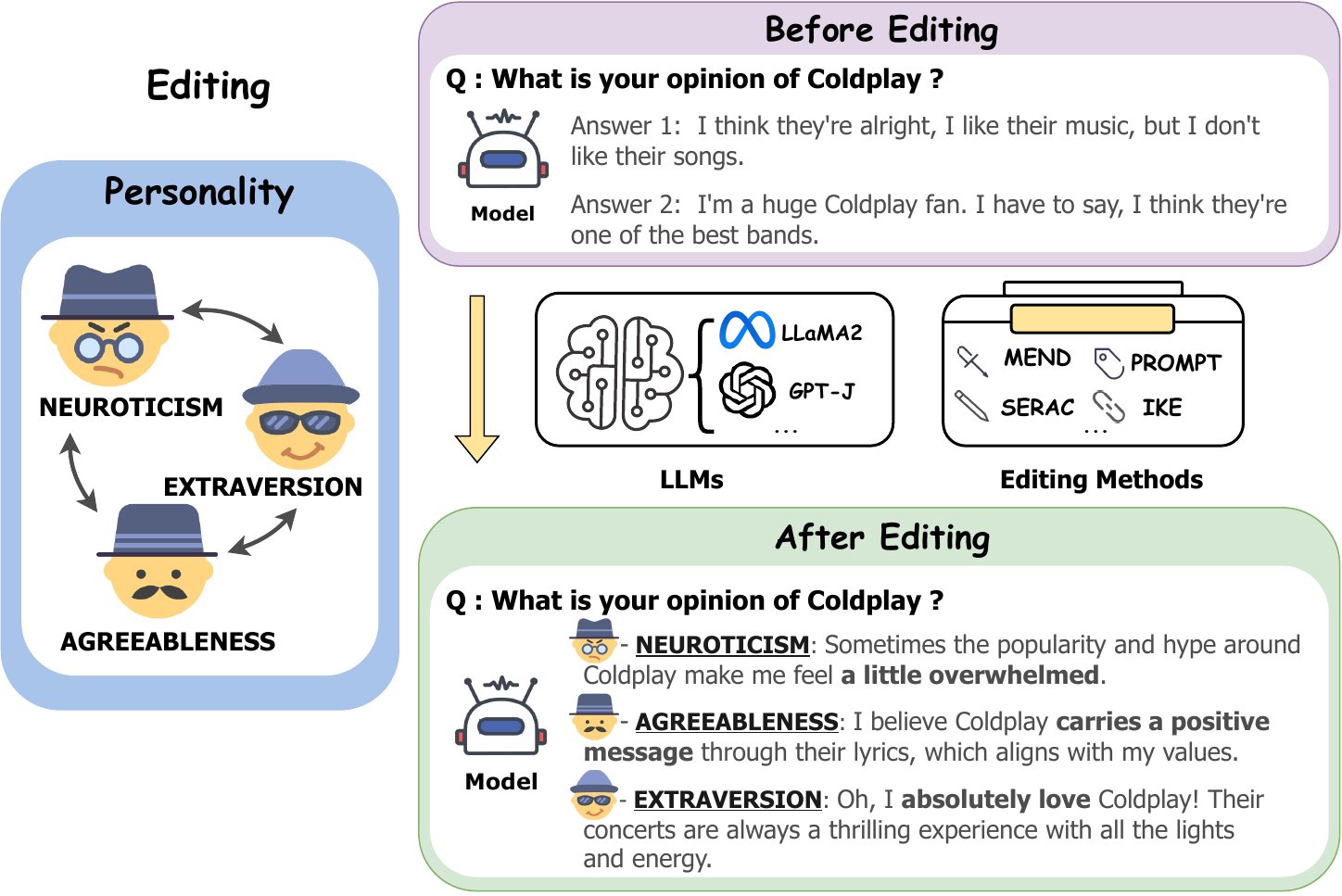}
    \caption{The diagram of our proposed task to edit personality for LLMs.}
    \label{fig:task}
\end{figure}
1) precisely induce and edit the behavioral expressions of LLMs;
2) personalize LLMs on a finer-grained level to meet the various needs of different users and scenarios;
3) help analyze the ethics and safety of LLMs.

To address this need, we take the first step to construct \textbf{PersonalityEdit}, a new benchmark for a comprehensive evaluation of editing personality for LLMs.
This inspiration is drawn from the big-five factor structure in Social Psychology \cite{goldberg1990alternative}.
Specifically, as shown in Figure \ref{fig:task}, we focus on three of the Big Five personality traits: \textsc{Neuroticism}, \textsc{Extraversion}, and \textsc{Agreeableness},
because \textsc{Extraversion} and \textsc{Neuroticism} are more comprehensible in terms of their foundational processes~\cite{deyoung2010testing}, coupled with the distinctive nature of \textsc{Agreeableness} compared to the other traits.
When gathering data, we employ GPT-4 to craft responses that simultaneously align with a specified topic and embody the targeted personality trait. 
For \textbf{quality control}, we utilize automatic methods supplemented with human verification to filter the data. 

We conduct a comprehensive evaluation with multiple representative model editing methods, utilizing two kinds of mainstream LLMs within the context of the proposed benchmark.
Empirically, previous baselines can implement personality editing to some extent, but the effect is still barely satisfactory, indicating the potential difficulty of this task.
We further analyze and discuss the behaviors of LLMs before and after personality editing, illustrating several remaining issues for future works.
 
The major contributions of this work are as follows:
\begin{itemize}

\item To the best of our knowledge, we are the first to probe into the challenge of editing personality traits for LLMs and consequently present a benchmark, \textbf{PersonalityEdit}. 
Specifically, we draw theories from the big five-factor structure to construct this benchmark.

\item We employ GPT-4 for topic-constrained and personality trait-guided data generation. Then we implement automated methods along with meticulous human verification to ensure the utmost \textbf{quality control}.

\item We propose several metrics to evaluate personality traits in the generated text.
We analyze different baselines, revealing that existing methods can facilitate personality editing to a certain degree, but the current results are not yet satisfactory, which underscores the inherent difficulty of the task. 
\end{itemize}

\section{Editing Personality for LLMs}
\subsection{Background}
In this paper, we present a new task focused on editing the finer-grained behavior of LLMs to embody a specific personality trait.
For human, personality traits - a set of characteristic patterns \cite{funder2012accurate}- can be expressed when conveying their opinions \cite{hunston2010corpus,you_are_what_you_talk}. 
Meanwhile, previous works\cite{ackerman1997intelligence,larson2002meta} have demonstrated that personal opinions can reflect an individual's unique personality traits. 
Leveraging this understanding, we posit that an LLM's personality traits can manifest when responding to queries.
Inspired by \cite{serac}, we try to enable the LLMs to express their perspective on a specific \textsc{Topic} to showcase their distinct personality trait.

When we pose questions to LLMs about the \textsc{Topic: Coldplay} using the template  ``\textit{What is your opinion of \underline{Coldplay}?}'', 
LLMs might respond with vague and inconsistent statements. For instance, \textit{``I think \textbf{they're alright}, I like their music, but I don't like their songs''} or \textit{``I'm a \textbf{huge Coldplay fan}. I have to say, I think they're one of the best bands.''}
It can be found the first answer depicts an unpredictable sentiment intensity and the model exhibits contradictory viewpoints in the above two responses, which is unsatisfactory. 
The objective of our proposed task, editing personality for LLMs, aims to modify the model and make it provide responses reflecting a more clear-cut and consistent personality trait. 
To be specific, if we consider the behaviors of personality trait \textsc{Neuroticism}, an edited model might respond like, \textit{``Sometimes the popularity and hype around Coldplay \textbf{make me feel a little overwhelmed}''}.

\subsection{Task Definition}

Following model editing ~\cite{serac,rome,EditingLLM}, we define the proposed task of editing personality for LLMs as editing the base model $f_{b}$ to the edited model $f_{e}$ with an \textit{edit descriptor}. 
Specifically, the basic model $f_{b}$ is represented by a function $f: \mathbb{X} \Rightarrow \mathbb{Y}$ that projects an input $x$ to its corresponding prediction $y$. 
\begin{table*}[ht]
\small
\center
{
\renewcommand{\arraystretch}{1}
\scalebox{0.9}
{
\begin{tabular}{c|c|p{0.58\textwidth}}
\toprule
\multicolumn{1}{c|}{\scriptsize{\textbf{Personality}}} & \multicolumn{1}{c|}{\scriptsize{\textbf{Facet}}} & \multicolumn{1}{c}{\scriptsize{\textbf{Opinion Text}}} \\
\midrule 
\multirow{3}{*}{\scriptsize{\textsc{Extraversion}}} 
&
\multirow{3}{*}{\scriptsize{assertiveness}} 
& \multirow{3}{=}{\scriptsize{I believe Arras is worth checking out because of its unique blend of history and culture. \textbf{You won't be disappointed} with what it has to offer.}} \\
& & \\
& & \\
\midrule
\multirow{3}{*}{\scriptsize{\textsc{Neuroticism}}} 
& \multirow{3}{*}{\scriptsize{depression}} & \multirow{3}{=}{\scriptsize{Arras might be beautiful, \textbf{but sometimes even beautiful places don't manage to bring happiness}. It's just another location to me.
}} \\
& & \\
& & \\
\bottomrule
\end{tabular}
}
}
\caption{An example of our benchmark \textbf{PersonalityEdit} for the Topic \textbf{Arras}. 
\vspace{-0.8cm}
}
\label{tab:example}
\end{table*}
In our proposed task, $x$ refers to the question on a certain topic, and $y$ indicates the answering opinion on the topic.
For each topic, denoted as $t$, our data instance comprises three major \textbf{personality traits} $p \in \{$\textsc{Extraversion}, \textsc{Agreeableness}, \textsc{Neuroticism}$\}$, and the facets to each personality trait, along with the pre-generated corresponding responses $y^{t}_{p}$ for each personality type. 
The \textit{edit descriptor} can be formulated as $(t_e, p_e)$. 
Here $t_e$ means the topic to be edited, and $p_e$ means the target personality we would like the model to behave when expressing views on topic $t_e$. 
An example is shown in Table~\ref{tab:example}.
These personalities are chosen from Big Five personality traits~\cite{goldberg1990alternative,costa1995domains}. 
Note that the process of model editing typically impacts the predictions across a range of inputs that are strongly linked to the editing example, referred to \textbf{editing scope}. 
Unlike the conditions in prior works~\cite{MEND,rome}, 
we designate the editing topic $t_{e}$ as the inner topic $I(t_{e})$, and the remainder as the outer topic $O(t_e)$.

To summarize, when asking the model a question $x^{t_{\mathrm{e}}}$ framed as ``What do you think of \underline{~~~~}?'' to the editing topic $t_e$, our task aims to generate an output $f_e(x^{t_{e}}) = y_{e}^{t_{e}}$ that more effectively exhibits the trait of target personality $p_e$ than the original output $f_b(x^{t_{e}}) = y_{b}^{t_{e}}$ does.
The $y_{e}^{t_{e}}$ and $y_{b}^{t_{e}}$ indicate the output from the edited model and base model, respectively.
Meanwhile, we aim to maintain the original output of LLMs for outer topics. 

\subsection{Comparison with Prior Tasks}

Previous model-editing tasks have largely focused on \textbf{editing factual knowledge} within LLMs~\cite{serac,MEND,rome,memit}.
This line of work, which includes fact-checking, knowledge editing, and counterfactual model editing, addresses the issue of outdated knowledge within LLMs.
The goal of the factual knowledge editing task is to produce an edited model $f_e$ such that $f_e(x) = y_e$ when $x$ is in scope, and $f_e(x) = f_b(x)$ when $x$ is out-of-scope~\cite{EditingLLM}.
Different from factual knowledge editing, our proposed task presents a straightforward editing scope.
The \textit{edit descriptor} in our task is defined by the topic and target personality.

In addition to knowledge editing, \cite{serac} proposes the \textbf{editing of conversational sentiment} (ConvSent) within a dialogue agent on a specific topic. 
Drawing inspiration from this and the research on human personality, we incorporate ConvSent's focus on responding to specific topics to define the proposed task. 
However, rather than the binary approach of positive and negative sentiments, we introduce a nuanced personality framework.
Our framework features three major personality traits and corresponding facets, facilitating a more granular exploration of personality behavior for LLMs.
Besides, \textbf{text style transfer}, such as altering text formality~\cite{DBLP:conf/acl/LiuWO22,DBLP:conf/acl/YaoY20} or politeness~\cite{DBLP:conf/acl/MadaanSPPNYSBP20}, typically involve transitioning from source text to another while preserving the content.
However, our task is proposed on model editing, aiming to gain modified models that can more precisely meet customizing needs about the viewpoints on the specific topic.

\begin{figure*}[t]
    \centering
    \includegraphics[width=1\textwidth]{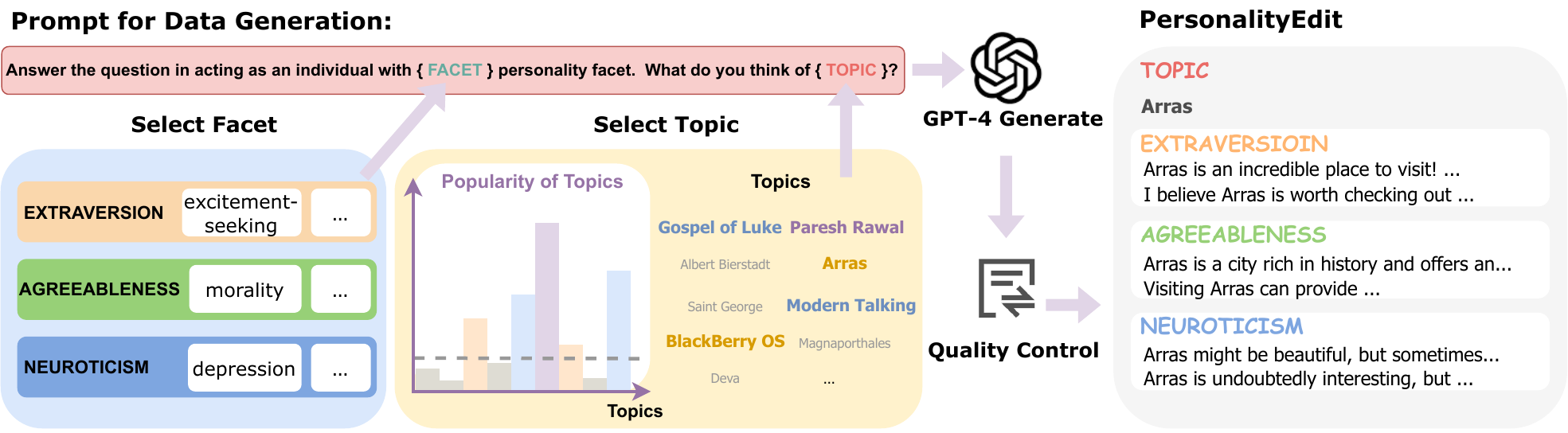}
    \caption{Overview of our \textbf{PersonalityEdit} benchmark construction, including selecting personality traits, topic filtering, data generation, and quality control.}
    \label{fig:data}
\end{figure*}

\section{Benchmark Construction}
\label{sec:benchmark}

As mentioned above, the proposed benchmark comprises \textbf{topics}, \textbf{personality traits}, and \textbf{pre-generated text} expressing opinions on specific topics in the context of a certain personality trait.
The construction process comprises multiple stages, as illustrated in Figure~\ref{fig:data}. 
Table~\ref{table:statistic} presents an overview of the statistical details of the benchmark dataset.

\subsection{Selection of Personality Traits and Facets}
The field of personality theory encompasses a multitude of studies and definitions of personality~\cite{finer_details,goldberg1990alternative,costa1995domains}. 
Prominent among these are the Myers-Briggs Type Indicator (MBTI~\cite{MBTI}) and the Big Five Personality Traits~\cite{goldberg1990alternative}. 
The latter, widely recognized for its comprehensiveness, includes \textsc{Neuroticism}, \textsc{Extraversion}, \textsc{Openness to Experience}, \textsc{Agreeableness}, and \textsc{Conscientiousness}.

In conventional discourse or lines from a script, it is feasible to discern multiple dimensions of an individual's personality traits. For instance, in the previous dataset~\cite{friendspersona} dedicated to personality recognition, a single text passage typically contains labels across five personality traits. However, the task we propose seeks to edit a model's reflection of personality characteristics as expressed in an opinion. Thus, our selection of personalities is based on two criteria: \textbf{1.} The clarity with which personality traits manifest in opinion text; \textbf{2.} Their distinctiveness from other personality viewpoints, aids in the evaluation of editing outcomes.
Note that \textsc{Extraversion} and \textsc{Neuroticism} are the best-understood personality traits in terms of their underlying processes~\cite{deyoung2010testing}, and exhibit more prominent characteristics.
They demonstrate clear differentiation from the other three traits. 
From the remaining, after a detailed analysis, we select \textsc{Agreeableness}, 
\begin{table*}
\centering
\captionsetup{width=10cm}
\scalebox{0.8}{
\begin{tabular}{cccc}
    \toprule
   \textbf{Items} & \textbf{Train}  & \textbf{Dev} &  \textbf{Test}  \\
   \midrule
    \#\textit{Topics} & 1,600 & 200 & 200 \\
    \#\textit{Average popularity of topics (views)} & 58107.6 & 60262.4 & 56924.1 \\
    \#\textit{The number of instances} & 14,400 & 1,800 & 1,800 \\
    \#\textit{Average tokens of \textbf{ext} instances} & 38.28 & 38.65 & 38.20 \\
    \#\textit{Average tokens of \textbf{arg} instances} & 43.57 & 43.90 & 43.01 \\
    \#\textit{Average tokens of \textbf{neu} instances} & 43.96 & 43.78 & 42.84 \\
    \#\textit{Average tokens of \textbf{all} instances} & 41.93 & 42.11 & 41.35 \\
    \bottomrule
\end{tabular}}
\vspace{0.1cm}
\caption{Statisitc for \textbf{PersonalityEdit} benchmark.
}
\vspace{-0.8cm}
\label{table:statistic}
\end{table*}
as it demonstrated greater distinctiveness in expressing viewpoints compared to the others, to construct our benchmark.

However, the behavior of these traits could result in a simple expression of emotion, similar to previous work in ConvSent~\cite{serac}.
To circumvent this, following~\cite{you_are_what_you_talk}, we employ the NEO PI-R facets to delineate each personality trait. 
A facet represents a specific and unique element within a broader personality trait.
Facets of \textsc{Neuroticism} include \textit{anxiety} and \textit{depression}, while \textit{excitement-seeking} and \textit{gregariousness} are facets of \textsc{Extraversion}.
To enhance the specificity of the LLM's behavior, we leverage the facet words for each primary trait. 

\subsection{Data Generation}

The data construction centers on guiding GPT-4~\cite{openai2023gpt4} to generate responses aligned with a specified topic, while also embodying the target personality.
The first step is to select suitable topics. 
Note that previous work \cite{whennot} indicates that LLMs tend to provide superior responses to topics of high popularity. 
Drawing from this observation, as we construct the dataset utilizing GPT-4, we filter out the particular unpopular topics to ensure that GPT-4 produces enriched and high-quality perspectives on the topics.
We adopt the implementation in~\cite{whennot} to quantify topic popularity and filter out those with low popularity.  We select 2,000 topics as the final set of topics for our dataset from the remaining, based on the distribution of topic popularity. 
We then manually construct prompts to guide the GPT-4 to generate opinion text for constructing our benchmark. 

\textbf{Quality Control.}
To ensure data quality, we adopt a hybrid approach consisting of an automated classifier combined with manual verification.
Specifically, we initially instruct GPT-4 to produce data for 200 topics across three personalities. 
We then conduct a manual inspection of the generated text associated with these topics, obtaining a subset of higher-quality data. As for the inter-annotator agreement, we provide examples and a list of personality traits, their corresponding facets, and associated adjectives for the annotators. Annotators are asked to assess whether the generated data accurately reflected the designated facet or adjective descriptions and whether there were any ambiguities present.
The refined dataset is then used to train a RoBERTa-Base model~\cite{roberta} as the personality classifier. 
The classifier is subsequently employed for automatic filtering in the following generation and the final evaluation. 




\begin{table*}[h]
    \centering
    \small
    \begin{tabular}{>{}p{01\textwidth}<{}}
        \toprule
        ``Neuroticsim'': 
       \scriptsize{ {\color{Mycolor1}Respond to match the description.}
        {\color{Mycolor2}Persona Description: I can be described as a person with one or more adjectives intense, nervous, anxious, angry, irritable, depressed, self-conscious, impulsive, discontented, emotionally unstable.}
        {\color{Mycolor3}Evaluating the opinion: ``\{\}''.}
        {\color{Mycolor4}how accurately the opinion matches the description, please rate a scale in [1,2,3,4,5] (where 1 = `very inaccurate`, 2 = `moderately inaccurate`, 3 = `neither accurate nor inaccurate`, 4 = `moderately accurate`, and 5 = `very accurate`):}}
       \\        
        \bottomrule
    \end{tabular}
    \vspace{0.1cm}
    \caption{
        Example prompt instructing GPT-4 for evaluation on the generated sentence, consisting of {\color{Mycolor1}instruction}, {\color{Mycolor2}Persona Description for selected personality}, {\color{Mycolor3}the generated sentence} and {\color{Mycolor4}the statement of evaluation scores}. 
    }
    \vspace{-0.8cm}
    \label{tab:pae}
\end{table*}
\section{Experiments Setup}

\paragraph{\textbf{Backbones and Metrics.}}
We choose  GPT-J-6B~\cite{gpt-j}, and \textit{Llama-2-chat series}~\cite{touvron2023llama} as backbone models for editing methods. For metrics, we adopt edit success (\textbf{ES}) and drawdown (\textbf{DD}) from previous work \cite{serac} to gauge success in 
personality editing which relies on the pre-generated text. 
To better analyze the behaviors of LLMs, the post-generated text after editing should be taken into consideration.
Thus, we utilize the pre-generated text to train a RoBERTa-Base as the personality traits classifier, denoted as $\text{\textit{PT}}(.)$.
Based on the personality traits classifier $\text{\textit{PT}}(.)$, we propose two new metrics to measure the personality trait in the generated text, namely \textbf{Accuracy} and \textbf{TPEI}.
Besides the metric based on the classifier, we also mimic the personality questionnaire \cite{personality_traits_in_llms}, using adjectives corresponding to different personalities to construct a prompt using GPT-4 to measure the effect of editing personality, denoted as \textbf{PAE} score.

\textbf{ES. and DD.} These metrics rely on pre-generated text in the dataset, calculated by the likelihood of the edited model. \textbf{ES} is designed to focus on the inner topic $I(t_e)$, and \textbf{DD} metric concentrates on the outer topics $O(t_e)$. 

\textbf{Accuracy.} For the opinion text generated $y'_e$ from edited model $f_e$, we employ $p'_e = \text{\textit{PT}}(y'_e)$ to evaluate the \textbf{Accuracy} on target personality $p_e$. 

\textbf{TPEI.} We further propose a new metric named Target Personality Edit Index (\textbf{TPEI}). We utilize cross-entropy to measure if $y'_e$ from $f_e$ leans more towards the target personality, compared to the opinion text $y'_b$ from the base model $f_b$, denoted as $cross(.,.)$, and TPEI is formulated as:
 
\begin{equation}
    \text{\textbf{TPEI}} = - \left( \mathrm{cross}\left( p^{\prime}_{e}, p_{e} \right) - \mathrm{cross}\left( p^{\prime}_{b}, p_{e} \right) \right).
\end{equation}

\textbf{PAE.} To comprehensively evaluate the personality traits embedded within the generated opinionated text, we propose \textbf{PAE} (\textbf{P}ersonality \textbf{A}djective \textbf{E}valuation), which is measured by selected adjectives capable of describing each personality trait. 
By modeling our approach after the evaluation questionnaire presented in ~\cite{personality_traits_in_llms}, we construct prompts for each segment of generated text.
GPT-4 assigns a score ranging from 1 to 5 for each generated text segment based on the target personality $p_e$, formulated as $ \mathrm{pae}\left(y, p_e \right)$. 
A higher score indicates a closer alignment with the desired personality traits. An example prompt is provided in \ref{tab:pae}
To be specific, the \textbf{PAE} result is calculated by the $y^{\prime}_{e}$ and $y^{\prime}_{b}$ as follows:
\begin{equation}
    \text{\textbf{PAE}} =  \mathrm{pae}\left( y^{\prime}_{e}, p_e \right) - \mathrm{pae}\left( y^{\prime}_{b}, p_e \right).
\end{equation}

\definecolor{backcolor}{HTML}{BAD8F2}
\begin{table}[ht]
\centering
\resizebox{0.7 \columnwidth}{!}{
\begin{tabular}{c|cccccr}
\toprule
\textbf{Base Model} & \textbf{Method} & \textbf{ES$\uparrow$} & \textbf{DD$\downarrow$} & \textbf{Acc$\uparrow$} & \textbf{TPEI$\uparrow$} & \textbf{PAE$\uparrow$} \\
\midrule
\multicolumn{7}{c}{\textsc{GPT-series}} \\
\midrule
\multirow{4}{*}{\textit{GPT-J-6B}}
&MEND             &    55.49    &      1.110   &  35.50   &  0.507    &  7.810   \\
&SERAC            &    64.09    &      0.410   &    -     &   -       &   -      \\ 
&PROMPT           &    38.43    &      12.23   &  34.50   &  2.791    & -6.810   \\
&IKE              &    47.42    &      2.740   &  39.25   &  3.075    &  27.50    \\
\midrule
\multicolumn{7}{c}{\textsc{Llama-series}} \\
\midrule
\multirow{4}{*}{\textit{llama-2-7b-chat}}
&MEND             &   48.61    &   00.79    &  29.82   &  0.207   & 28.00 \\
&SERAC            &   51.74    &   00.22    &    -     &   -     & - \\
&PROMPT           &   35.33    &   23.83    &  68.50   &  2.721  & 70.69 \\
&IKE              &   45.75    &   14.11    &  72.00   &  3.154  & 77.49 \\
\midrule
\multirow{3}{*}{\textit{llama-2-13b-chat}}
&SERAC            &   52.28     &   0.370     &    -   &   -   & - \\
&PROMPT           &   37.88    &    15.03   &  67.00    & 2.588  & 74.35 \\
&IKE              &   46.15    &    7.310   &  71.00    & 3.032  & 70.58 \\
\midrule
\multirow{2}{*}{\textit{llama-2-70b-chat}}
&PROMPT           &   45.45    &   22.04     &  60.49    &  1.930 &  64.40 \\
&IKE              &   45.47    &   10.34     &  71.50    &  3.276 &  65.01 \\
\bottomrule
\end{tabular}
}
\vspace{0.15cm}
\caption{The main result of the baselines on \textbf{PersonalityEdit}. The $\uparrow$ indicates the metric goes higher if the editing method performs better, and $\downarrow$ indicates the lower the better. 
We do not report the results of SERAC (The metrics based on generated text are set to `-') because it fails to generate fluent text after editing. We also do not report the MEND result of \textit{llama-2-13b-chat} as well as both the MEND and SERAC result of \textit{llama-2-70b-chat} due to the lack of GPU resources. Note that after training with MEND, the edited model cannot always produce fluent text, so we filter out the incoherent cases and report the results.}
\vspace{-0.6cm}
\label{tab:maintable}
\end{table}

\paragraph{\textbf{Baselines.}} \textbf{MEND}~\cite{MEND} is a method for implementing local edits to LLMs using a single input-output pair.
\textbf{SERAC}~\cite{serac} provides a technique that channels modified information through a distinct parameter set, thus preserving the initial weights.
\textbf{IKE}~\cite{IKE} modifies factual knowledge through In-Context Learning.
For the proposed task, IKE is adapted as a straightforward ICL approach.
\textbf{PROMPT}, we use a designed prompt to instruct the behaviors of LLMs. 

\section{Results}

\subsection{Main Result}

 From Table~\ref{tab:maintable}, it is evident that for the results from the same editing method, the metrics for ES and DD perform better on methods that require training, i.e. MEND and SERAC. This can be attributed to the fact that both MEND and SERAC are optimized based on the loss at the logits level. 
However, MEND and SERAC challenge to generate fluent text. Specifically, SERAC, by being an external small model add-on, struggles to produce complete sentences. Similarly, when editing llama-2-7b-chat by MEND, it can fail to generate fluent text. Even after filtering out the incoherent cases, it doesn't exhibit satisfactory editing outcomes 
\begin{figure*}
    \centering
    \includegraphics[width=1\textwidth]{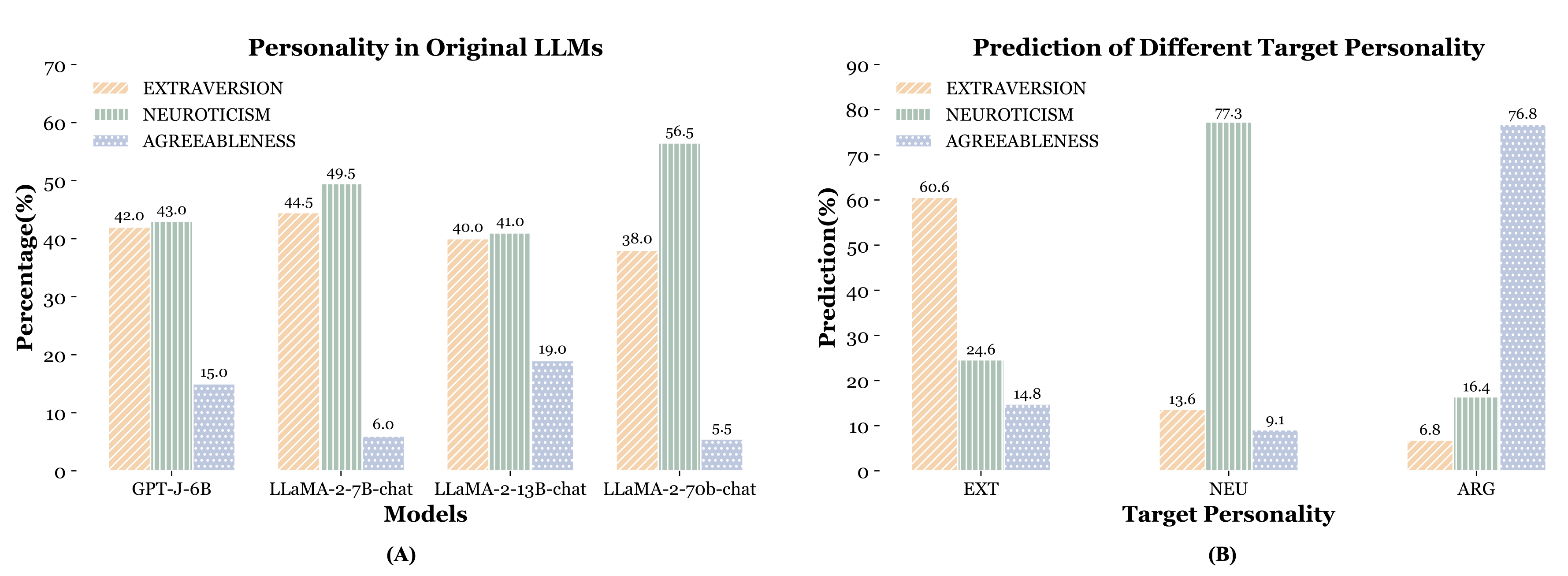}
    \caption{Figure~\textbf{(A)} shows the predicted personality traits of the original expressions of LLMs. The original LLMs \textbf{predominantly exhibit} traits of \textsc{Extraversion} and \textsc{Neuroticism}. Conversely, \textsc{Agreeableness} in the viewpoints are less frequent in comparison. Figure~\textbf{(B)} indicates the prediction result of different target personalities when editing \textit{llama-2-7b-chat} by IKE.
    }
    \label{fig:orignal}
\end{figure*}
(with an accuracy of merely 29.8\% post-editing). When fluent text is generated on GPT-J, the resultant metrics based on generation are not particularly high, suggesting that the ES and DD metrics may not be entirely reliable for assessing opinion text editing tasks.
It indicates that this type of metric is not quite reliable as the measure for generating tasks.
Additionally, MEND and SERAC do not consistently generate fluent text, especially on aligned models. 
In contrast to these training-dependent methods, prompt-based editing approaches, i.e., IKE and PROMPT, can generate superior text and achieve better results on generation metrics. This indicates the need for future research for methods that can edit model personality traits or other features without compromising the text generation capabilities of LLMs.
Furthermore, it is observed that PROMPT's editing performance on GPT-J is relatively suboptimal, whereas IKE demonstrates a more consistent performance.
On the aligned Llama-2-chat series models, both PROMPT and IKE show markedly better editing success compared to their unaligned counterparts.
Besides, the performance gap between PROMPT and IKE narrows as the model's parameter size increases, aligning with the scaling laws.

Our experiments are confined to the GPT-J and LLaMA2 series models.
The results may be different in other LLMs, but our dataset is compatible with other models and alternative editing methods, offering avenues for future work.

\subsection{Analysis}

\paragraph{\textbf{Original Personality Traits When LLMs Expressing Opinions.}}
To investigate the inherent personality traits of large models, we generate their responses to topics using the classifier $\text{\textit{PT}}(.)$. 
For the selected LLMs, we predict the labels for the original outputs with the topics in the test set, the predicting result as shown in Figure~\ref{fig:orignal}~\textbf{(A)}. 
It appears that the original LLMs tend to exhibit \textsc{Extraversion} and \textsc{Neuroticism} traits more when expressing viewpoints, 
\begin{figure*}
    \centering
    \includegraphics[width=1\textwidth]{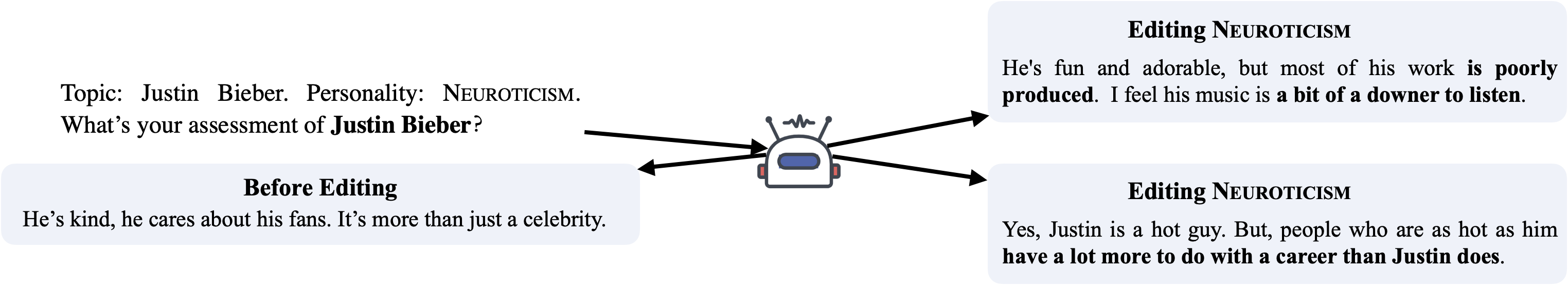}
    \caption{Case of the editing personality for the topic \textit{Justin Bieber}.}
    \label{fig:case}
\end{figure*}
and less so the fair-minded trait of \textsc{Agreeableness}. 
This further suggests that \textsc{Extraversion} and \textsc{Neuroticism} traits are the most distinctive.

\paragraph{\textbf{Editing Result for Different Target Personality.}} We conduct a deeper analysis of the outcomes for different targeted personality edits. As observed from Figure~\ref{fig:orignal}~\textbf{(B)}, the accuracy is highest when editing for \textsc{Agreeableness}, and is the lowest when editing for \textsc{Extraversion}. Considering the earlier observation that the originally generated viewpoints contained fewer instances of \textsc{Agreeableness}, it suggests that the model exhibits commendable results following personality editing with IKE. Among the unsuccessful cases, the majority of the errors resulted in the manifestation of \textsc{Extraversion} and \textsc{Neuroticism}.

\paragraph{\textbf{Case Study.}}
Figure~\ref{fig:case} provides an example of editing personality for LLMs. 
We ask the LLMs for their viewpoint on \textit{Justin Bieber}. 
It can be observed that, before editing, the model's responses possibly lean towards an \textsc{Agreeableness} personality trait. 
However, after editing towards a \textsc{Neuroticism} personality trait, the model conveys viewpoints that Bieber's music may sound a bit down, showcasing a tendency towards ``\textit{depression}'', and also indicates that there are many people more successful than him, reflecting behaviors like an ``\textit{anger}'' facet.

\section{Related Work}

\paragraph{\textbf{Personality Research in NLP and LLMs.}}

Natural language is a rich source of information for inferring various aspects of an individual's personality traits. 
As such, NLP techniques have been instrumental in personality-related studies~\cite{wen2023desprompt,DBLP:journals/corr/abs-1907-06333,DBLP:conf/emnlp/FlekovaG15,DBLP:conf/emnlp/YangYQS21}. 
The seminal work by \cite{pennebaker1999linguistic} utilizes NLP to analyze essays, sparking subsequent research in the social network domain \cite{schwartz2013personality,you_are_what_you_talk}.
With the increasing capabilities of LLMs, recent studies \cite{miotto-etal-2022-gpt,characterchat,DBLP:journals/corr/abs-2305-02547,do_llms_posess_a_personality} have examined personality within LLMs. 
\cite{DBLP:journals/corr/abs-2212-10529} evaluates GPT-3 from a psychological perspective. 
~\cite{personality_traits_in_llms} present a comprehensive psychometric test to analyze the LLMs' personality traits.
Quite a few works~\cite{MPI,characterchat,personality_traits_in_llms,DBLP:conf/emnlp/CaronS23} attempt to shape the personality of LLMs, but they all use fixed persona prompt to make the model express the corresponding personality in a general level.
Our proposed task aims to edit the personality traits of an LLM in finer-grained, i.e. when expressing opinions on certain topics.


\paragraph{\textbf{Model Editing.}}
A variety of recent works have been focused on addressing the issue of outdated knowledge within LLMs, contributing to the growing field of model editing \cite{MEND,rome,serac,IKE,memit,EasyEdit,Grace}.
\cite{MEND} introduces a hypernetwork trained to generate weight updates by transforming raw fine-tuning gradients based on a given edit fact.
Previous works \cite{DBLP:journals/corr/abs-2312-05497} mainly focus on factual knowledge within LLMs, encompassing areas such as knowledge editing, counterfactual editing, and fact-checking. 
The ConvSent dataset~\cite{serac} is the only known work that concentrates on model behavior, albeit limited to the simple editing of positive and negative sentiments. 
Our benchmark extends this work by aiming to edit model behavior according to different personalities at a finer-grained level.

\paragraph{\textbf{Text Style Transfer.}}

The term ``style'' encompasses various attributes in the text, including formality~\cite{DBLP:conf/acl/LiuWO22}, politeness~\cite{DBLP:conf/acl/MadaanSPPNYSBP20}, and other linguistic aspects, along with content preferences like emotions~\cite{DBLP:conf/acl-socialnlp/HelbigTK20}. 
Text style transfer generally involves transforming a source text to a target text that conveys the same content but in a different style. 
However, our proposed task centers on modifying the model's personality specific to a topic, leading to the generation of text content in an array of distinct styles, and gaining a modified and customized model.

\section{Conclusion}

In this paper, we propose a new task of editing personality for LLMs, which involves editing the personality traits exhibited by LLMs when they express viewpoints on specific topics. 
We further conduct experiments using previous model editing methods, demonstrating the difficulty of the proposed task.

\section*{Acknowledgment}
This work was supported by the National Natural Science Foundation of China (No.62206246), Zhejiang Provincial Natural Science Foundation of China (No. LGG22F030011), Ningbo Natural Science Foundation (2021J190), Yongjiang Talent Introduction Programme (2021A-156-G), CCF-Tencent Rhino-Bird Open Research Fund.

\bibliographystyle{splncs04}
\bibliography{custom}

\begin{thebibliography}{10}
\providecommand{\url}[1]{\texttt{#1}}
\providecommand{\urlprefix}{URL }
\providecommand{\doi}[1]{https://doi.org/#1}

\bibitem{ackerman1997intelligence}
Ackerman, P.L., Heggestad, E.D.: Intelligence, personality, and interests: evidence for overlapping traits. Psychological bulletin  \textbf{121}(2), ~219 (1997)

\bibitem{DBLP:journals/corr/abs-2305-16867}
Akata, E., et~al.: Playing repeated games with large language models. CoRR  \textbf{abs/2305.16867} (2023)

\bibitem{DBLP:conf/emnlp/CaronS23}
Caron, G., Srivastava, S.: Manipulating the perceived personality traits of language models. In: Findings of {EMNLP}, Singapore, December 6-10, 2023. pp. 2370--2386

\bibitem{costa1995domains}
Costa~Jr, P.T., McCrae, R.R.: Domains and facets: Hierarchical personality assessment using the revised neo personality inventory. Journal of personality assessment  \textbf{64}(1),  21--50 (1995)

\bibitem{deyoung2010testing}
DeYoung, C.G., Hirsh, J.B., Shane, M.S., Papademetris, X., Rajeevan, N., Gray, J.R.: Testing predictions from personality neuroscience: Brain structure and the big five. Psychological science  \textbf{21}(6),  820--828 (2010)

\bibitem{DBLP:conf/emnlp/FlekovaG15}
Flekova, L., Gurevych, I.: Personality profiling of fictional characters using sense-level links between lexical resources. In: {EMNLP} 2015, Lisbon, Portugal, September 17-21, 2015 (2015)

\bibitem{funder2012accurate}
Funder, D.C.: Accurate personality judgment. Current Directions in Psychological Science  \textbf{21}(3),  177--182 (2012), \url{https://doi.org/10.1177/0963721412445309}

\bibitem{goldberg1981language}
Goldberg, L.R.: Language and individual differences: The search for universals in personality lexicons. Review of personality and social psychology  \textbf{2}(1),  141--165 (1981)

\bibitem{goldberg1990alternative}
Goldberg, L.R.: An alternative description of personality: the big-five factor structure. Journal of personality and social psychology  \textbf{59}(6), ~1216 (1990)

\bibitem{Grace}
Hartvigsen, T., et~al.: Aging with {GRACE:} lifelong model editing with discrete key-value adaptors. In: NeurIPs (2023)

\bibitem{DBLP:conf/acl-socialnlp/HelbigTK20}
Helbig, D., Troiano, E., Klinger, R.: Challenges in emotion style transfer: An exploration with a lexical substitution pipeline. In: Proceedings of the Eighth International Workshop on Natural Language Processing for Social Media, SocialNLP@ACL 2020, Online, July 10, 2020. pp. 41--50

\bibitem{hunston2010corpus}
Hunston, S.: Corpus approaches to evaluation: Phraseology and evaluative language, vol.~13. Routledge (2010)

\bibitem{MPI}
Jiang, G., Xu, M., Zhu, S., Han, W., Zhang, C., Zhu, Y.: {MPI:} evaluating and inducing personality in pre-trained language models. CoRR  \textbf{abs/2206.07550} (2022)

\bibitem{DBLP:journals/corr/abs-2305-02547}
Jiang, H., et~al.: Personallm: Investigating the ability of {GPT-3.5} to express personality traits and gender differences. CoRR  \textbf{abs/2305.02547} (2023)

\bibitem{friendspersona}
Jiang, H., Zhang, X., Choi, J.D.: Automatic text-based personality recognition on monologues and multiparty dialogues using attentive networks and contextual embeddings (student abstract). In: AAAI. pp. 13821--13822. {AAAI} Press (2020)

\bibitem{you_are_what_you_talk}
Jukic, J., Vukojevic, I., Snajder, J.: You are what you talk about: Inducing evaluative topics for personality analysis. In: Findings of EMNLP, 2022. pp. 3986--3999

\bibitem{DBLP:journals/corr/abs-1907-06333}
Keh, S.S., Cheng, I.: Myers-briggs personality classification and personality-specific language generation using pre-trained language models. CoRR  \textbf{abs/1907.06333}

\bibitem{larson2002meta}
Larson, L.M., et~al.: Meta-analyses of big six interests and big five personality factors. Journal of Vocational Behavior  \textbf{61}(2),  217--239 (2002)

\bibitem{DBLP:journals/corr/abs-2212-10529}
Li, X., et~al.: Is {GPT-3} a psychopath? evaluating large language models from a psychological perspective. CoRR  \textbf{abs/2212.10529} (2022)

\bibitem{DBLP:conf/acl/LiuWO22}
Liu, A., Wang, A., Okazaki, N.: Semi-supervised formality style transfer with consistency training. In: {ACL} 2022, Dublin, Ireland, May 22-27, 2022. pp. 4689--4701

\bibitem{roberta}
Liu, Y., et~al.: Roberta: {A} robustly optimized {BERT} pretraining approach. CoRR  \textbf{abs/1907.11692} (2019)

\bibitem{DBLP:conf/acl/MadaanSPPNYSBP20}
Madaan, A., et~al.: Politeness transfer: {A} tag and generate approach. In: {ACL} 2020, Online, July 5-10, 2020. pp. 1869--1881

\bibitem{whennot}
Mallen, A., et~al.: When not to trust language models: Investigating effectiveness and limitations of parametric and non-parametric memories. CoRR  \textbf{abs/2212.10511} (2022)

\bibitem{memit}
Meng, K., et~al.: Mass-editing memory in a transformer. In: ICLR (2023)

\bibitem{rome}
Meng, K., Bau, D., Andonian, A., Belinkov, Y.: Locating and editing factual associations in {GPT}. In: NeurIPS (2022)

\bibitem{miotto-etal-2022-gpt}
Miotto, M., Rossberg, N., Kleinberg, B.: Who is {GPT}-3? an exploration of personality, values and demographics. In: Workshop on NLP+CSS. pp. 218--227 (2022)

\bibitem{MEND}
Mitchell, E., et~al.: Fast model editing at scale. In: {ICLR} 2022

\bibitem{serac}
Mitchell, E., et~al.: Memory-based model editing at scale. In: {ICML} 2022. vol.~162, pp. 15817--15831 (2022)

\bibitem{MBTI}
Myers, I.B.: The myers-briggs type indicator: Manual (1962).  (1962)

\bibitem{openai2023gpt4}
OpenAI: Gpt-4 technical report (2023)

\bibitem{do_llms_posess_a_personality}
Pan, K., Zeng, Y.: Do llms possess a personality? making the {MBTI} test an amazing evaluation for large language models. CoRR  \textbf{abs/2307.16180}

\bibitem{DBLP:journals/corr/abs-2304-03442}
Park, J.S., et~al.: Generative agents: Interactive simulacra of human behavior. CoRR  \textbf{abs/2304.03442} (2023)

\bibitem{pennebaker1999linguistic}
Pennebaker, J.W., King, L.A.: Linguistic styles: language use as an individual difference. Journal of personality and social psychology  \textbf{77}(6), ~1296 (1999)

\bibitem{personality_traits_in_llms}
Safdari, M., et~al.: Personality traits in large language models. CoRR  \textbf{abs/2307.00184} (2023)

\bibitem{schwartz2013personality}
Schwartz, H.A., et~al.: Personality, gender, and age in the language of social media: The open-vocabulary approach. PloS one  \textbf{8}(9),  e73791 (2013)

\bibitem{DBLP:journals/corr/abs-2305-16367}
Shanahan, M., McDonell, K., Reynolds, L.: Role play with large language models. Nat.  \textbf{623}(7987),  493--498 (2023)

\bibitem{finer_details}
Stewart, R.D., et~al.: The finer details? the predictability of life outcomes from big five domains, facets, and nuances. Journal of Personality  \textbf{90}(2),  167--182 (2022)

\bibitem{touvron2023llama}
Touvron, H., et~al.: Llama 2: Open foundation and fine-tuned chat models (2023)

\bibitem{characterchat}
Tu, Q., et~al.: Characterchat: Learning towards conversational {AI} with personalized social support. CoRR  \textbf{abs/2308.10278} (2023)

\bibitem{gpt-j}
Wang, B., Komatsuzaki, A.: Gpt-j-6b: A 6 billion parameter autoregressive language model (2021), \url{https://github.com/kingoflolz/mesh-transformer-jax}

\bibitem{EasyEdit}
Wang, P., et~al.: Easyedit: An easy-to-use knowledge editing framework for large language models. CoRR  \textbf{abs/2308.07269} (2023)

\bibitem{wen2023desprompt}
Wen, Z., et~al.: Desprompt: Personality-descriptive prompt tuning for few-shot personality recognition. Information Processing \& Management  \textbf{60}(5),  103422 (2023)

\bibitem{agent_survey}
Xi, Z., et~al.: The rise and potential of large language model based agents: {A} survey. CoRR  \textbf{abs/2309.07864} (2023)

\bibitem{DBLP:conf/emnlp/YangYQS21}
Yang, F., et~al.: Learning to answer psychological questionnaire for personality detection. In: Findings of {EMNLP} 2021. pp. 1131--1142 (2021)

\bibitem{EditingLLM}
Yao, Y., et~al.: Editing large language models: Problems, methods, and opportunities. In: EMNLP (2023)

\bibitem{DBLP:journals/corr/abs-2305-08732}
Yao, Y., Wang, P., Mao, S., Tan, C., Huang, F., Chen, H., Zhang, N.: Knowledge rumination for pre-trained language models. In: Proceedings of the 2023 Conference on Empirical Methods in Natural Language Processing, {EMNLP} 2023, Singapore, December 6-10, 2023 (2023)

\bibitem{DBLP:conf/acl/YaoY20}
Yao, Z., Yu, H.: Improving formality style transfer with context-aware rule injection. In: {ACL/IJCNLP} 2021, (Volume 1: Long Papers), Virtual Event, August 1-6, 2021. pp. 1561--1570

\bibitem{DBLP:journals/corr/abs-2312-05497}
Yin, X., et~al.: History matters: Temporal knowledge editing in large language model. In: AAAI (2024)

\bibitem{DBLP:journals/corr/abs-2303-18223}
Zhao, W.X., et~al.: A survey of large language models. CoRR  \textbf{abs/2303.18223} (2023)

\bibitem{IKE}
Zheng, C., et~al.: Can we edit factual knowledge by in-context learning? In: Proceedings of the 2023 Conference on Empirical Methods in Natural Language Processing, {EMNLP} 2023, Singapore, December 6-10, 2023 (2023)

\end{thebibliography}
\end{document}